\documentclass[a4paper]{article}
\usepackage{graphicx}
\usepackage{twocolceurws}
\usepackage{amsmath}
\usepackage{caption}
\usepackage{graphicx,wrapfig}

\title{Real-time Claim Detection from News Articles and Retrieval of Semantically-Similar Factchecks}
\author{Ben Adler  \\ 
                London UK  \\ ben@thelogically.co.uk
\and
Giacomo Boscaini-Gilroy \\
                London UK \\ giacomo@logically.co.uk
}

\institution{Logically}

\begin{document}
\maketitle

\begin{abstract}
Factchecking has always been a part of the journalistic process. However with newsroom budgets shrinking \cite{Pew16} it is coming under increasing pressure just as the amount of false information circulating is on the rise \cite{Martens18}. We therefore propose a method to increase the efficiency of the factchecking process, using the latest developments in Natural Language Processing (NLP). This method allows us to compare incoming claims to an existing corpus and return similar, factchecked, claims in a live system---allowing  factcheckers to work simultaneously without duplicating their work. 
\end{abstract}

\section{Introduction}

In recent years, the spread of misinformation has become a growing concern for researchers and the public at large \cite{Martens18}. Researchers at MIT found that social media users are more likely to share false information than true information \cite{Vosoughi2018}. Due to renewed focus on finding ways to foster healthy political conversation, the profile of factcheckers has been raised.
\par
Factcheckers positively influence public debate by publishing good quality information and asking politicians and journalists to retract misleading or false statements. By calling out lies and the blurring of the truth, they make those in positions of power accountable. This is a result of labour intensive work that involves monitoring the news for spurious claims and carrying out rigorous research to judge credibility. So far, it has only been possible to scale their output upwards by hiring more personnel. This is problematic because newsrooms need significant resources to employ factcheckers. Publication budgets have been decreasing, resulting in a steady decline in the size of their workforce \cite{Pew16}. Factchecking is not a directly profitable activity, which negatively affects the allocation of resources towards it in for-profit organisations. It is often taken on by charities and philanthropists instead.
\par
To compensate for this shortfall, our strategy is to harness the latest developments in NLP to make factchecking more efficient and therefore less costly. To this end, the new field of automated factchecking has captured the imagination of both non-profits and start-ups \cite{Graves18, Babakar16, Thorne18}. It aims to speed up certain aspects of the factchecking process rather than create AI that can replace factchecking personnel. This includes monitoring claims that are made in the news, aiding decisions about which statements are the most important to check and automatically retrieving existing factchecks that are relevant to a new claim.
\par
The claim detection and claim clustering methods that we set out in this paper can be applied to each of these. We sought to devise a system that would automatically detect claims in articles and compare them to previously submitted claims. Storing the results to allow a factchecker's work on one of these claims to be easily transferred to others in the same cluster.

\section{Claim Detection}

\subsection{Related Work}

It is important to decide what sentences are claims before attempting to cluster them. The first such claim detection system to have been created is ClaimBuster \cite{Hassan2017}, which scores sentences with an SVM to determine how likely they are to be politically pertinent statements. Similarly, ClaimRank \cite{Jaradat2018} uses real claims checked by factchecking institutions as training data in order to surface sentences that are worthy of factchecking.
\par
These methods deal with the question of what is a politically interesting claim. In order to classify the objective qualities of what set apart different types of claims, the ClaimBuster team created PolitiTax \cite{Caraballo18}, a taxonomy of claims, and factchecking organisation Full Fact \cite{Konstantinovskiy18} developed their preferred annotation schema for statements in consultation with their own factcheckers. This research provides a more solid framework within which to construct claim detection classifiers.
\par
 The above considers whether or not a sentence is a claim, but often claims are subsections of sentences and multiple claims might be found in one sentence. In order to accommodate this, \cite{Levy2017} proposes extracting phrases called Context Dependent Claims (CDC) that are relevant to a certain `Topic'. Along these lines, \cite{Arslan19} proposes new definitions for frames to be incorporated into FrameNet \cite{FrameNet} that are specific to facts, in particular those found in a political context.

\subsection{Method}

It is much easier to build a dataset and reliably evaluate a model if the starting definitions are clear and objective. Questions around what is an interesting or pertinent claim are inherently subjective. For example, it is obvious that a politician will judge their opponents' claims to be more important to factcheck than their own.
\par
Therefore, we built on the methodologies that dealt with the objective qualities of claims, which were the PolitiTax and Full Fact taxonomies. We annotated sentences from our own database of news articles based on a combination of these. We also used the Full Fact definition of a \textbf{claim} as \textit{a statement about the world that can be checked}. Some examples of claims according to this definition are shown in Table \ref{table:exampleclaim}. We decided the first statement was a claim since it declares the occurrence of an event, while the second was considered not to be a claim as it is an expression of feeling.

\begin{table}[ht]
\begin{center}
\caption{Examples of claims taken from real articles.}
\label{table:exampleclaim}
\begin{tabular}{|l|l|l|l|}

\hline
Sentence & Claim? \\

\hline
In its 2015 order, the NGT had banned & Yes \\ the plying of petrol vehicles older than & \\ 15 years and diesel vehicles older than & \\ 10 years in the National Capital Region & \\ (NCR). & \\

\hline
In my view, farmers should not just & No \\ rely on agriculture but also adopt & \\ dairy farming. & \\

\hline

\end{tabular}
\end{center}
\end{table}

Full Fact's approach centred around using sentence embeddings as a feature engineering step, followed by a simple classifier such as logistic regression, which is what we used. They used Facebook's sentence embeddings, InferSent \cite{infersent}, which was a recent breakthrough at the time. Such is the speed of new development in the field that since then, several papers describing textual embeddings have been published. Due to the fact that we had already evaluated embeddings for clustering, and therefore knew our system would rely on Google USE Large \cite{Cer18}, we decided to use this instead. We compared this to TFIDF and Full Fact's results as baselines. The results are displayed in Table \ref{table:detection}.
\par
However, ClaimBuster and Full Fact focused on live factchecking of TV debates. Logically is a news aggregator and we analyse the bodies of published news stories. We found that in our corpus, the majority of sentences are claims and therefore our model needed to be as selective as possible. In practice, we choose to filter out sentences that are predictions since generally the substance of the claim cannot be fully checked until after the event has occurred. Likewise, we try to remove claims based on personal experience or anecdotal evidence as they are difficult to verify.

\begin{table}[ht]
\begin{center}
\caption{Claim Detection Results.}
\label{table:detection}
\begin{tabular}{|l|l|l|l|}
\hline
Embedding Method & P & R & F1\\
\hline

Google USE Large& 0.90   &   0.89  &    0.89                                                        \\
\cite{Cer18} &&& \\
Full Fact (not on       & 0.88         & 0.80                        & 0.83                                                         \\
the same data) \cite{Konstantinovskiy18}       & & &                                                      \\
TFIDF (Baseline)        & 0.84 & 0.84 & 0.84                  \\
\cite{TFIDF} &&& \\
\hline

\end{tabular}
\end{center}
\end{table}

\section{Claim Clustering}

\subsection{Related Work}

Traditional text clustering methods, using TFIDF and some clustering algorithm, are poorly suited to the problem of clustering and comparing short texts, as they can be semantically very similar but use different words. This is a manifestation of the  the data sparsity problem with Bag-of-Words (BoW) models. \cite{Song15}. Dimensionality reduction methods such as Latent Dirichlet Allocation (LDA) can help solve this problem by giving a dense approximation of this sparse representation \cite{Blei03}. More recently, efforts in this area have used text embedding-based systems in order to capture dense representation of the texts \cite{Wangetal2015}. Much of this recent work has relied on the increase of focus in word and text embeddings. Text embeddings have been an increasingly popular tool in NLP since the introduction of Word2Vec \cite{Mikolov2013}, and since then the number of different embeddings has exploded. While many focus on giving a vector representation of a word, an increasing number now exist that will give a vector representation of a entire sentence or text. Following on from this work, we seek to devise a system that can run online, performing text clustering on the embeddings of texts one at a time

\subsubsection{Text Embeddings}

Some considerations to bear in mind when deciding on an embedding scheme to use are: the size of the final vector, the complexity of the model itself and, if using a pretrained implementation, the data the model has been trained on and whether it is trained in a supervised or unsupervised manner. 
\par
The size of the embedding can have numerous results downstream. In our example we will be doing distance calculations on the resultant vectors and therefore any increase in length will increase the complexity of those distance calculations. We would therefore like as short a vector as possible, but we still wish to capture all salient information about the claim; longer vectors have more capacity to store information, both salient and non-salient. 
\par
A similar effect is seen for the complexity of the model. A more complicated model, with more trainable parameters, may be able to capture finer details about the text, but it will require a larger corpus to achieve this, and will require more computational time to calculate the embeddings. We should therefore attempt to find the simplest embedding system that can accurately solve our problem. 
\par
When attempting to use pretrained models to help in other areas, it is always important to ensure that the models you are using are trained on similar material, to increase the chance that their findings will generalise to the new problem. Many unsupervised text embeddings are trained on the CommonCrawl \footnote{CommonCrawl found at http://commoncrawl.org/} dataset of approx. 840 billion tokens. This gives a huge amount of data across many domains, but requires a similarly huge amount of computing power to train on the entire dataset. Supervised datasets are unlikely ever to approach such scale as they require human annotations which can be expensive to assemble. The SNLI entailment dataset is an example of a large open source dataset \cite{snli:emnlp2015}. It features pairs of sentences along with labels specifying whether or not one entails the other. Google's Universal Sentence Encoder (USE) \cite{Cer18} is a sentence embedding created with a hybrid supervised/unsupervised method, leveraging both the vast amounts of unsupervised training data and the extra detail that can be derived from a supervised method. The SNLI dataset and the related MultiNLI dataset are often used for this because textual entailment is seen as a good basis for general Natural Language Understanding (NLU) \cite{Williams18}.

\subsection{Choosing an embedding}
\par
In order to choose an embedding, we sought a dataset to represent our problem. Although no perfect matches exist, we decided upon the Quora duplicate question dataset \cite{Quora} as the best match. To study the embeddings, we computed the euclidean distance between the two questions using various embeddings, to study the distance between semantically similar and dissimilar questions. 

\begin{figure}[h]
    \centering
    \includegraphics[width=\linewidth]{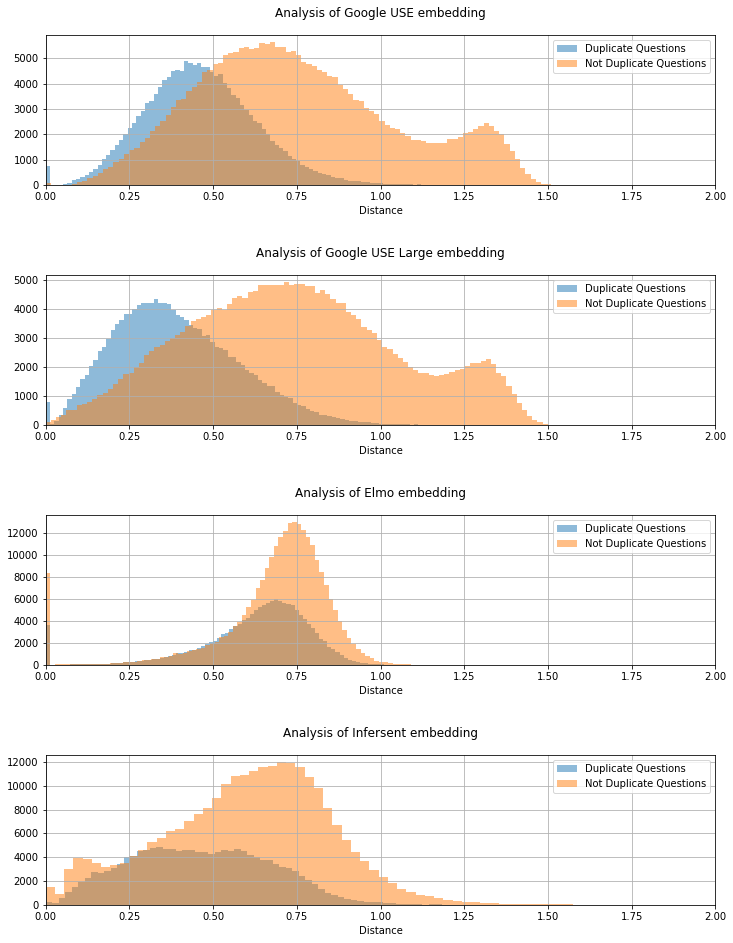}
    \caption{Analysis of Different Embeddings on the Quora Question Answering Dataset}
\end{figure}

\begin{table*}[ht]
\begin{center}
\caption{Comparing Sentence Embeddings for Clustering News Claims.}
\label{table:clustering}

\bigskip

\begin{tabular}{|l|lllll|}
\hline
Embedding & Time & Number & Number & Percentage of   & Percentage of\\
method & taken (s) & of claims & of clusters & claims in& claims in clusters  \\
&&clustered&& majority clusters & of one story \\
\hline
Elmo \cite{elmo}            & 122.87         & 156                        & 21                 & 57.05\%                                   & 3.84\%                                         \\
Googe USE \cite{Cer18}        & 117.16         & 926                        & 46                 & 57.95\%                                   & 4.21\%                                      \\
Google USE Large \cite{Cer18}& 95.06          & 726                        & 63                 & 60.74\%                                   & 7.02\%                                          \\
Infersent \cite{infersent}       & 623.00         & 260                        & 34                 & 63.08\%                                   & 10.0\%                                         \\
TFIDF (Baseline) \cite{TFIDF}          & 25.97          & 533                        & 58                 & 62.85\%                                   & 7.12\% \\
\hline

\end{tabular}
\end{center}
\end{table*}



\par
The graphs in figure 1 show the distances between duplicate and non-duplicate questions using different embedding systems. The X axis shows the euclidean distance between vectors and the Y axis frequency.  A perfect result would be a blue peak to the left and an entirely disconnected orange spike to the right, showing that all non-duplicate questions have a greater euclidean distance than the least similar duplicate pair of questions. As can be clearly seen in the figure above, Elmo \cite{elmo} and Infersent \cite{infersent} show almost no separation and therefore cannot be considered good models for this problem. A much greater disparity is shown by the Google USE models \cite{Cer18}, and even more for the Google USE Large model. In fact the Google USE Large achieved a F1 score of 0.71 for this task without any specific training, simply by choosing a threshold below which all sentence pairs are considered duplicates.

\par
In order to test whether these results generalised to our domain, we devised a test that would make use of what little data we had to evaluate. We had no original data on whether sentences were semantically similar, but we did have a corpus of articles clustered into stories. Working on the assumption that similar claims would be more likely to be in the same story, we developed an equation to judge how well our corpus of sentences was clustered, rewarding clustering which matches the article clustering and the total number of claims clustered. The precise formula is given below, where $P_{os}$ is the proportion of claims in clusters from one story cluster, $P_{cc}$ is the proportion of claims in the correct claim cluster, where they are from the most common story cluster, and $N_{c}$ is the number of claims placed in clusters. A,B and C are parameters to tune.

\begin{align*}
  \left( A \times P_{os} + B \times  P_{cc} \vphantom{\sum} \right) \times   \left(C \times N_{c} \right)
\end{align*}

\begingroup\vspace*{-\baselineskip}
\captionof{figure}{Formula to assess the correctness of claim clusters based on article clusters}
\vspace*{\baselineskip}\endgroup

\par
This method is limited in how well it can represent the problem, but it can give indications as to a good or bad clustering method or embedding, and can act as a check that the findings we obtained from the Quora dataset will generalise to our domain. We ran code which vectorized ~2,000 sentences and then used the DBScan clustering method \cite{DBScan} to cluster using a grid search to find the best $\epsilon$ value, maximizing this formula. We used DBScan as it mirrored the clustering method used to derive the original article clusters. The results for this experiment can be found in Table \ref{table:clustering}. We included TFIDF in the experiment as a baseline to judge other results. It is not suitable for our eventual purposes, but it the basis of the original keyword-based model used to build the clusters \footnote{Described in the newslens paper \cite{Laban17}}. That being said, TFIDF performs very well, with only Google USE Large and Infersent coming close in terms of `accuracy'. In the case of Infersent, this comes with the penalty of a much smaller number of claims included in the clusters. Google USE Large, however, clusters a greater number and for this reason we chose to use Google's USE Large. \footnote{Google USE Large is the Transformer based model, found at https://tfhub.dev/google/universal-sentence-encoder-large/3, whereas Google USE uses a DAN architecture}

\par
Since Google USE Large was the best-performing embedding in both the tests we devised, this was our chosen embedding to use for clustering. However as can be seen from the results shown above, this is not a perfect solution and the inaccuracy here will introduce inaccuracy further down the clustering pipeline.

\subsection{Clustering Method}
We decided to follow a methodology upon the DBScan method of clustering \cite{DBScan}. DBScan considers all distances between pairs of points. If they are under $\epsilon$ then those two are linked. Once the number of connected points exceeds a minimum size threshold, they are considered a cluster and all other points are considered to be unclustered. This method is advantageous for our purposes because unlike other methods, such as K-Means, it does not require the number of clusters to be specified. To create a system that can build clusters dynamically, adding one point at a time, we set the minimum cluster size to one, meaning that every point is a member of a cluster. 
\par
A potential disadvantage of this method is that because points require only one connection to a cluster to join it, they may only be related to one point in the cluster, but be considered in the same cluster as all of them. In small examples this is not a problem as all points in the cluster should be very similar. However as the number of points being considered grows, this behaviour raises the prospect of one or several borderline clustering decisions leading to massive clusters made from tenuous connections between genuine clusters. To mitigate this problem we used a method described in the Newslens paper \cite{Laban17} to solve a similar problem when clustering entire articles. We stored all of our claims in a graph with the connections between them added when the distance between them was determined to be less than $\epsilon$. To determine the final clusters we run a Louvain Community Detection \cite{Louvain} over this graph to split it into defined communities. This improved the compactness of a cluster. When clustering claims one by one, this algorithm can be performed on the connected subgraph featuring the new claim, to reduce the computation required.
\par
As this method involves distance calculations between the claim being added and every existing claim, the time taken to add one claim will increase roughly linearly with respect to the number of previous claims. Through much optimization we have brought the computational time down to approximately 300ms per claim, which stays fairly static with respect to the number of previous claims.

\section{Next Steps}
The clustering described above is heavily dependent on the embedding used. The rate of advances in this field has been rapid in recent years, but an embedding will always be an imperfect representation of an claim and therefore always an area of improvement. A domain specific-embedding will likely offer a more accurate representation but creates problems with clustering claims from different domains. They also require a huge amount of data to give a good model and that is not possible in all domains.

\subsubsection*{Acknowledgements}

Thanks to Anil Bandhakavi, Tom Dakin and Felicity Handley for their time, advice and proofreading.

\bibliographystyle{alpha} 
\bibliography{ref}

\end{document}